\documentclass{article}
\usepackage{amssymb}
\usepackage{amsmath}
\usepackage{graphicx}
\usepackage{wrapfig}
\usepackage{bm}
\usepackage{multirow}
\usepackage[table]{xcolor}
\usepackage{tikzducks}

\usepackage{multirow}
\usepackage{adjustbox}
\usepackage{caption}
\usepackage[preprint]{corl_2025} 
\usepackage{booktabs}

\usepackage{subcaption}

\title{
Hand-Eye Autonomous Delivery: Learning Humanoid Navigation, Locomotion and Reaching}

\author{
  Sirui Chen$^*$\\
  Stanford University\\
  United States \\
  \texttt{ericcsr@stanford.edu} \\
  \And
  Yufei Ye$^*$ \\
  Stanford University\\
  United States \\
  \texttt{yufeiy2@stanford.edu} \\
  \And
  Zi-ang Cao$^*$ \\
  Stanford University\\
  United States \\
  \texttt{ziangcao@stanford.edu} \\
  \AND
  Jennifer Lew \\
  Stanford University\\
  United States \\
  \texttt{jennlew@stanford.edu} \\
  \And
  Pei Xu \\
  Stanford University\\
  United States \\
  \texttt{peixu@stanford.edu} \\
  \And
  C. Karen Liu \\
  Stanford University\\
  United States \\
  \texttt{karenliu@cs.stanford.edu} \\
}

\newcommand*{\ours}[0]{HEAD}

\begin{document}
\maketitle


\begin{abstract}
We propose Hand-Eye Autonomous Delivery (\ours), a framework that learns navigation, locomotion, and reaching skills for humanoids, directly from human motion and vision perception data. We take a modular approach where the high-level planner commands the target position and orientation of the hands and eyes of the humanoid, delivered by the low-level policy that controls the whole-body movements. Specifically, the low-level whole-body controller learns to track the three points (eyes, left hand, and right hand) from existing large-scale human motion capture data while high-level policy learns from human data collected by Aria glasses. Our modular approach decouples the ego-centric vision perception from physical actions, promoting efficient learning and scalability to novel scenes. We evaluate our method both in simulation and in the real-world, demonstrating humanoid's capabilities to navigate and reach in complex environments designed for humans.

\end{abstract}

\section{Introduction}
\label{sec:intro}



\begin{wrapfigure}{r}{0.62\textwidth}
  \begin{center}
  \includegraphics[width=\linewidth]{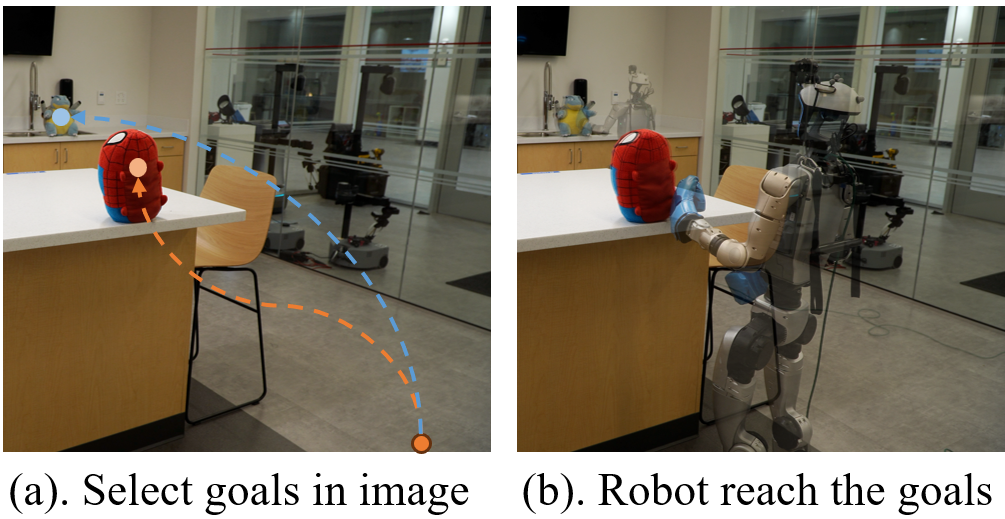}
  \end{center}
  \vspace{-1em}
  \caption{Given a user selected goal from the humanoid's egocentric view, the humanoid is able to navigate in the 3D world and reach the goal.}
  \vspace{-1em}
\end{wrapfigure}
Any human manipulation task begins with moving close to the target object so we can perceive it and touch it. Consequently, a fundamental skill a humanoid must master is the ability to deliver its end-effectors and cameras to the place they are needed in a 3D environments designed by and for humans. One possible approach is to directly combine existing navigation methods for mobile robots and reaching methods for manipulators. 
However, this often leads to control strategies that either spatially isolate upper-body manipulation from lower-body locomotion, or a lack of coordination necessary for seamless transition between navigation and reaching behavior.


We propose a hand-eye autonomous delivery (\ours) system for humanoids, designed to fully utilize their human-like morphology to achieve concurrent navigation, locomotion and reaching tasks in a coordinated manner. While learning from human demonstrations is a promising avenue, training all three skills end-to-end would require heterogeneous human data involving both egocentric vision and full-body motion. Instead, we adopt a modular approach that decouples egocentric perception from physical actions, enabling flexible training of whole body navigation, locomotion and reaching using different sources of human data and different algorithms. This design also mitigates the challenges of training a unified visuomotor policy. Our framework consists of a high-level policy that predicts target positions and orientations for the humanoid’s eyes and hands, and a low-level controller that executes the corresponding full-body motions.

Given a command to reach for and touch an object, indicated by a point in the initial RGB image perceived by the humanoid, the high-level policy predicts head positions and orientations to guide the humanoid toward the target while keeping it in view and navigating around obstacles. Once the target is within arm’s reach, the high-level policy also controls the hands to make contact with the object. Existing visual navigation methods \cite{levine2023learning,navila} often abstract the robot as a point mass, limiting actions to 2D movements on the ground plane. While suitable for wheeled robots, such assumptions are insufficient for humanoids, which must coordinate an articulated body to navigate complex 3D spaces, simultaneously reaching for and avoiding objects at varying heights. To enable this 3D navigation capability, our method leverages a mix of different datasets for different purposes---internet large-scale human exploration datasets for generalization to new scenes, mid-scale demonstrations in the target environment for mitigating domain shift due to perception, and a small amount of robot-specific experience for mitigating domain shift due to embodiment gap.

The low-level whole-body controller is trained to track three key points—the eyes, left hand, and right hand—using large-scale human motion capture data. We employ imitation-based reinforcement learning (RL) for training, leveraging the diversity of large datasets to handle a wide range of target configurations. Training such a whole-body policy with imitation-based RL presents three major challenges. First, unlike full-body tracking, our targets are spatially sparse, guiding only three points. Second, whole-body skills require the upper and lower body to perform different tasks simultaneously, necessitating a large number of demonstrations to cover the joint action space. Third, obtaining accurate root position and velocity information in the real world is difficult, requiring a more robust policy that functions without precise root data. We address the first challenge by formulating a GAN-based RL framework that imitates the distribution of human demonstrations, rather than relying on specific full-body trajectories as policy input. To tackle the second challenge, we design two separate discriminators to reward the upper and lower body independently, promoting composability and coordination between them. Finally, to address the third challenge, we train a policy that does not depend on root position or velocity in world coordinates; instead, global information is inferred from the navigation goal and estimated via the onboard camera.

We evaluate each component of our system separately to better understand their contributions. For the low-level policy, we assess its ability to accurately track diverse target points across a wide range of motions. For the high-level navigation module, we find that using both human and robot data is essential for achieving reliable performance, while large-scale human data significantly improves generalization to novel environments. Finally, we integrate the full system and deploy it on a humanoid (Unitree G1) in the real world, demonstrating robust navigation and reaching performance in a human indoor space. 

\section{Related Works}
\label{sec:related}
\textbf{Learning from Human Data.}
Following the growing popularity of humanoid robots, using human data to train humanoids has started to attract attention. Internet-scale human videos provide abundant training sources, making them suitable for pre-training implicit visual representation~\cite{r3m,dasari2023unbiased,o2024open}. But their embodiment and observation gap make them less efficient and less relevant to learn a specific skill. Recent works show that high-quality task-relevant human data can benefit  robot training, for both tabletop manipulation \cite{egomimic, dexcap,umi}, and indoor navigation \cite{navila}. 
Recent works propose creative ways to collect these data with various focuses. Portable devices, such as VR headsets \cite{arcap}, AR glasses \cite{ariaglasses}, or SLAM cameras \cite{dexcap}, can capture multiple modalities of human data, including head and hand poses, which can be easily transferred to the robot. While different tasks typically require different forms of human data for effective learning, we advocate for modular system interfaced with 3-point tracking for joint navigation, reaching, and locomotion. 

\textbf{Humanoid whole body control.}
Recent advances in humanoid hardware have made humanoids more accessible for academic research. To enable a humanoid to achieve meaningful tasks, the whole body controller (WBC) serves as a cornerstone in balancing the humanoid robot and coordinating whole-body movement. Traditionally, optimal control-based whole controller\cite{crocoddyl, atlas, sreenath2011compliant} has enabled a humanoid to walk, jump, and locomote through challenging terrain given a detailed kinematic trajectory to track. However, such high-quality kinematic trajectories are hard to obtain and highly specific to particular robot kinematics. Recently, reinforcement learning (RL) based whole body controller has shown impressive result to directly learn from human data\cite{deepmimic, phc, humanplus, h2o, exbody2}. Among them, most of WBCs are designed to track human joint positions\cite{deepmimic, phc, humanplus, h2o}, which requires human whole body pose as input. To utilize sparse input that are easier to capture from virtual reality (VR) devices, \cite{omnih2o, hover, questsim} build their WBC to track human head and hand positions which can be accurately obtained from off-the-shelf VR headsets. Alternatively, \cite{lu2024mobile} use VR headset and pedals to control upper-body and lower-body seperately. We use head and wrists tracking similar to \cite{hover, omnih2o} as our WBC interface as it allows directly transfer human data in the task space. Compared to prior methods, our WBC also track head and wrists orientation which allow more versitile manipulation skills such as twisting the wrists.

\textbf{Navigation.}
Extensive research in visual navigation has largely treated the robot as a point mass operating in a 2D plane. In long-term navigation,  prior-work uses different exploration strategies ranging from local method~\cite{kuipers1991robot,bourgault2002information}, global method~\cite{thrun1999minerva,yamauchi1997frontier,holz2011comparative} to end-to-end learning of goal-driven policies~\cite{chaplot2018gated,mnih2015human,sridhar2024nomad}, planning over floor-plan waypoints or semantic landmarks to achieve robust performance across large spaces. Short-term navigation similarly relies on this 2D abstraction but focuses on socially compliant and reactive behaviors—dynamic obstacle avoidance and human–robot interaction \cite{hirose2023sacson,mumm2011human,mavrogiannis2023core,mavrogiannis2023core}. 
The closest work on humanoid platforms is NaVila~\cite{navila}, which applies long-horizon 2D waypoint navigation to a bipedal robot but decouples perception from locomotion and ignores full-body reaching. By comparison, our approach studies short-term 3D navigation directly through whole-body control, and to mitigate the embodiment gap between human and humanoid. 



\section{Method}
\label{sec:method}
\begin{figure}[h]
  \centering
  \includegraphics[width=\linewidth]{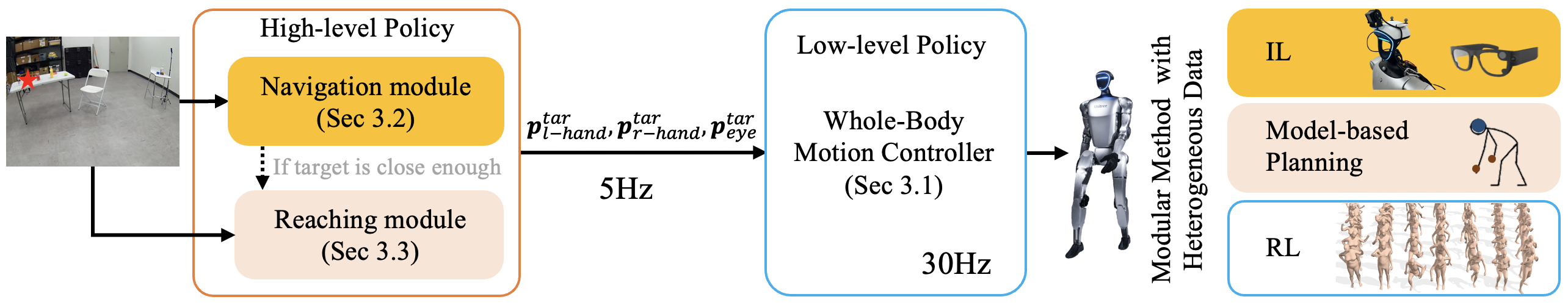}
  \caption{\textbf{System overview}: \ours~consists of a high-level policy with two modules, navigation and reaching, and a low-level policy that coordinates the whole-body motion. The high-level policy provides hand-eye tracking targets at a lower frequency while the whole-body controller tracks the hand-eye targets at a higher frequency. The learning-based navigation module learns from a mixed training dataset to map RGB ego-vision perception to camera target trajectories. The model-based reaching module generates hand-eye target poses. The low-level whole-body controller is trained using imitation-based RL on a set of human motion capture data.}
  \label{fig:overview}
\end{figure}

Given a selected point on the initial RGB image observed by the robot, our system, \ours, enables the humanoid to reach that point in the physical 3D world using its hand. \ours~is a modular system composed of a high-level policy for navigation and reaching, and a low-level policy for whole-body control (Figure~\ref{fig:overview}). The core idea is that both navigation and reaching can be accomplished by commanding the same low-level whole-body policy to track the 6D poses of the head and hands.


\subsection{Whole-body Controller}
\begin{figure}
    \centering
    \includegraphics[width=\linewidth]{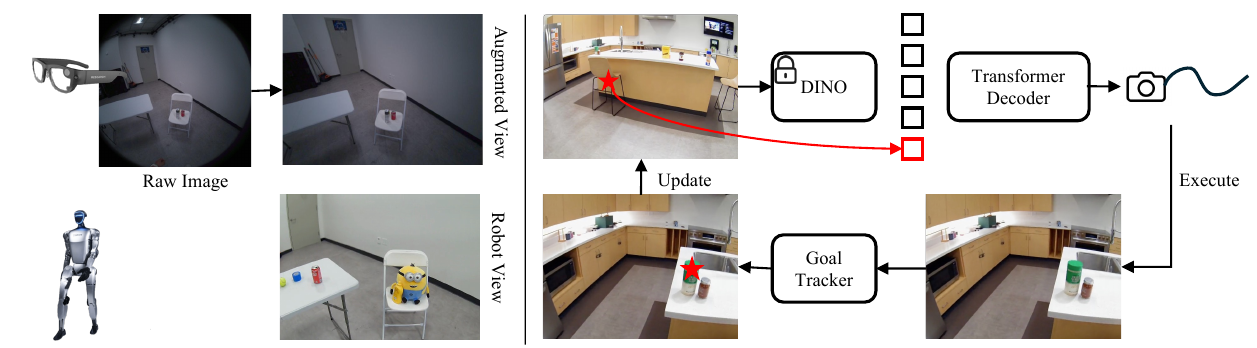}
    \caption{\textbf{Navigation Training Data} (left): we augment images (undistortion and homography transform) collected from Aria Glasses to make them resemble robot views. \textbf{Navigation Module Overview} (right):  given an image and a goal as 2D point during inference, we extract DINO features, append the goal coordinate, and feed them to a transformer decoder to predict the future eye (camera) trajectory. The low-level whole-body controller executes the prediction and obtains a new observation. The goal is then tracked in the new image using an off-the-shelf point tracker.}
    \label{fig:nav_method}
    \vspace{-20pt}
\end{figure}


Given the target hand-eye positions and orientations provided by the high-level policy,
the low-level, whole-body controller controls the humanoid through PD servos.
To let the humanoid behave in a human-like manner while tracking arbitrary targets,
we train the control policy through a GAN-like method~\cite{iccgan} to perform motion imitation from unstructured motion data under the framework RL, combined with goal-directed control for target position and orientation tracking.
Unlike two-stage distillation methods~\cite{human_humanoid,omnih2o},
our method trains the low-level control policy in an end-to-end way for real-world deployment.

\textbf{Curating Human Motion Dataset.} We find that the quality of motion retargeting significantly impacts policy performance. 
We curated a 5-hour dataset by retargeting human motion capture data to the G1 robot, from the AMASS \cite{amass} and OMOMO \cite{omomo} datasets. The retargeting is achieved using keypoint matching similar to \cite{h2o}. The collected motions ensemble representative behaviors across both manipulation and locomotion domains.
Dataset will be open source upon acceptance.

\textbf{Deployable Observation Space.} To support real-world deployment, the observation space must be restricted to information accessible from the robot’s onboard sensors. 
Our observation vector 
consists of robot link poses $\bm{p_\text{link}}$ 
in the robot's local coordinate frame in two consecutive time steps and joint velocity $\bm{\dot{q}}$ locally.
It does \emph{not} include any future information or rely on any privileged data in the world coordinate frame, such as root position and linear velocity, which are difficult to obtain outside of simulation. 
We found that removing the dependency on privileged information outperforms any alternative approaches that depend on reconstructed or predicted substitues. 

\textbf{Motion Imitation.} We decouple the full-body motion into upper and lower body groups, and employ two discriminators simultaneously to perform imitation learning. By doing so, the policy can learn the combination of the poses from upper and lower body parts, instead of being limited by fixed full-body poses provided in the motion dataset.
The GAN-like approach of RL allows the policy to imitate motions from arbitrary segments in the motion dataset without needing to generate or obtain a full trajectory of imitation beforehand, while fulfilling the tracking task.

\textbf{Sparse Target Tracking.}
To avoid introducing target information defined in the global space,
we represent the tracking target, as the input to the policy network, through relative transformations: $\bm{g} = [\bm{p}^\text{tar}_\text{eye}\ominus\bm{p}_\text{eye};\bm{p}^\text{tar}_\text{l-hand}\ominus\bm{p}_\text{l-hand};\bm{p}^\text{tar}_\text{r-hand}\ominus\bm{p}_\text{r-hand}]$ where $\ominus$ denotes the relative transformation operator, and “tar” refers to the target pose. To perform tracking, during training,
we define the goal-directed reward based on $\bm{g}$ after the action is executed at each timestep.

\textbf{Sim-to-real Considerations.}
Along with the task reward of target tracking,
we additionally define
a regularization term 
to aid sim-to-real transfer. To further improve robustness, we apply extensive domain randomization over dynamics parameters and sensor noise during training. 

We employ the multi-objective learning framework from~\cite{composite} to perform policy training, 
while optimizing the two imitation objectives using rewards provided by the discriminators and the goal-directed objective through the manually defined reward function at the same time.
We refer to the supplementary materials for the implementation details.


\subsection{Navigation Module}
\label{sec:nav}
Given a low-level whole-body controller capable of tracking three points, our navigation module guides the robot to a designated goal specified as a 2D point in the initial RGB image observed by the robot. During inference, the navigation model takes the current RGB image from the navigation camera along with the tracked 2D goal—provided by a point tracker~\cite{karaev2024cotracker3}—and predicts the future eye trajectory in both position and orientation (Fig.~\ref{fig:nav_method} right).
Specifically, we extract DINO features of the input image $I_t$ and add positional embedding to the goal $g_t$. We pass them to a transformer decoder to output a future camera trajectory $C_{t:t+T}$ as transformations relative to the previous frame. 

\textbf{Collecting Human Data.} We propose an automatic method to use Aria Glasses for collecting goal-conditioned human training data as tuples of images, future camera trajectories, and 2D goals ($I_t, C_{t:t+T}, g_t$).
The glasses provide accurate camera poses, static point clouds and gaze estimation for all captured data. We approximate the current goal by finding the closest point in the static point cloud along the future gaze vector and projecting it onto the image plane via current camera pose. 

\textbf{Domain Shift.} However, a navigation model trained on a limited set of human data struggles with two potential domain shifts that need to be addressed. First, to improve generalization to unseen scenes, we incorporate the large-scale egocentric dataset Aria Digital Twin (ADT)~\cite{adt}, which contains 400-minute of various indoor activities such as cleaning and cooking. Thanks to our automatic data curation pipeline, we can easily convert any Aria Glasses data into goal-conditioned navigation training data. Second, due to the embodiement gap between the robot and an average human adult, there is a significant disparity in visual perception. To align the Aria Glasses’ wide fisheye view with the robot’s narrower camera, we apply undistortion and homography transforms to produce virtual views of robot from human data (Fig.~\ref{fig:nav_method} left). (See appendix for details.) Beyond visual discrepancies, humans and robots also operate at different speeds. Empirically, we find that the robot moves approximately 7× slower than humans, so we subsample robot videos accordingly during training.

We also collect a small amount of robot data by commanding it to navigate while recording its head poses with the mocap system. 
We co-train the navigation module with both human and robot data. 


\begin{wrapfigure}{r}{0.45\textwidth}
  \centering
  \includegraphics[width=\linewidth]{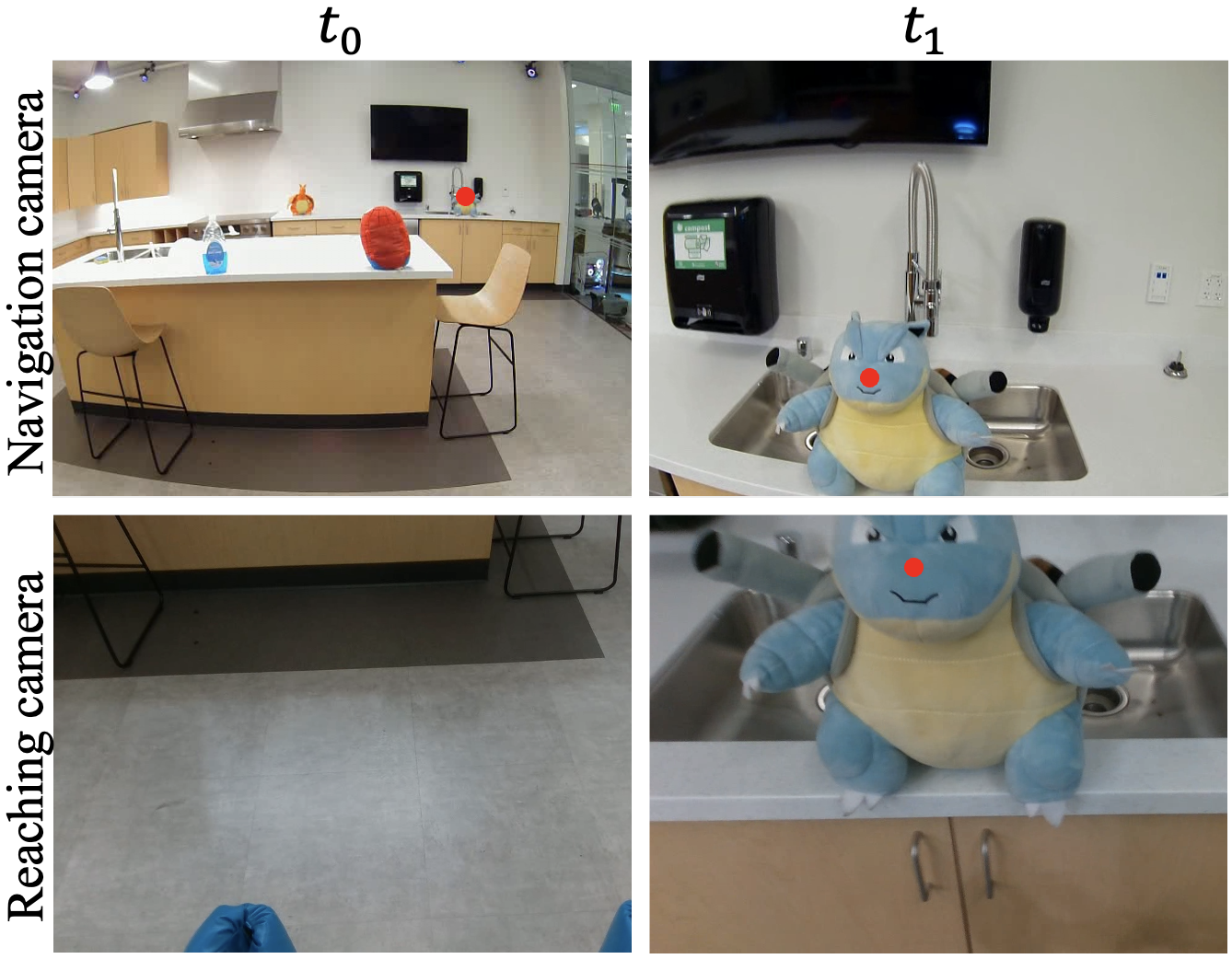}
  \caption{Robot is away from the goal at $t_0$, goal is only visible in navigation camera. At transition time $t_1$, the goal is visible by the reach camera and close enough to the robot.}
    \label{fig:transition}
    \vspace{-20pt}
\end{wrapfigure}

\subsection{Reaching Module}
While the navigation module drives the robot toward the target object at room scale, the reaching module handles the final approach to touch the object. We use a second downward-facing RGB-D camera with a narrower, zoomed-in FoV for reaching.
 The high-level policy switches from navigation to reaching while the low-level policy continues running to ensure a smooth transition.
 
\textbf{Navigation-Reaching Transition.}
 The navigation policy hands control to the reaching module when the target object enters the downward-looking RGB-D camera’s view and is within reaching range. The goal is transferred to the RGB-D frame via correspondence matching \cite{loftr}.
 
\textbf{Reaching the Target.}
 The reaching module approximates the goal as a 3D hand position and computes the target hand orientation and head 6D pose for the low-level policy. Since the tracked goal only specifies the hand position, we solve Inverse Kinematics (IK) using Mink \cite{mink} to infer missing head poses and hand orientations. To ensure a smooth high-level transition and keep the robot posture looking natural, we initialize the IK optimization from the current robot state and add an objective term to encourage small change in the center of mass position and pelvis orientation.

\section{Experiment}
We first specify our hardware setup (\ref{sec:hard_exp}). We evaluate our overall hand-eye delivery system with novel objects in different environments(Sec.~\ref{sec:overall_exp}). Then, we analyze individual modules -- the design choice of the whole-body controller (Sec.~\ref{sec:wbc_exp}), and contribution of each data ingredient to train the navigation module (Sec.~\ref{sec:nav_exp}). 

\subsection{Hardware Setup}
\label{sec:hard_exp}
\begin{figure}
  \centering
  \includegraphics[width=\linewidth]{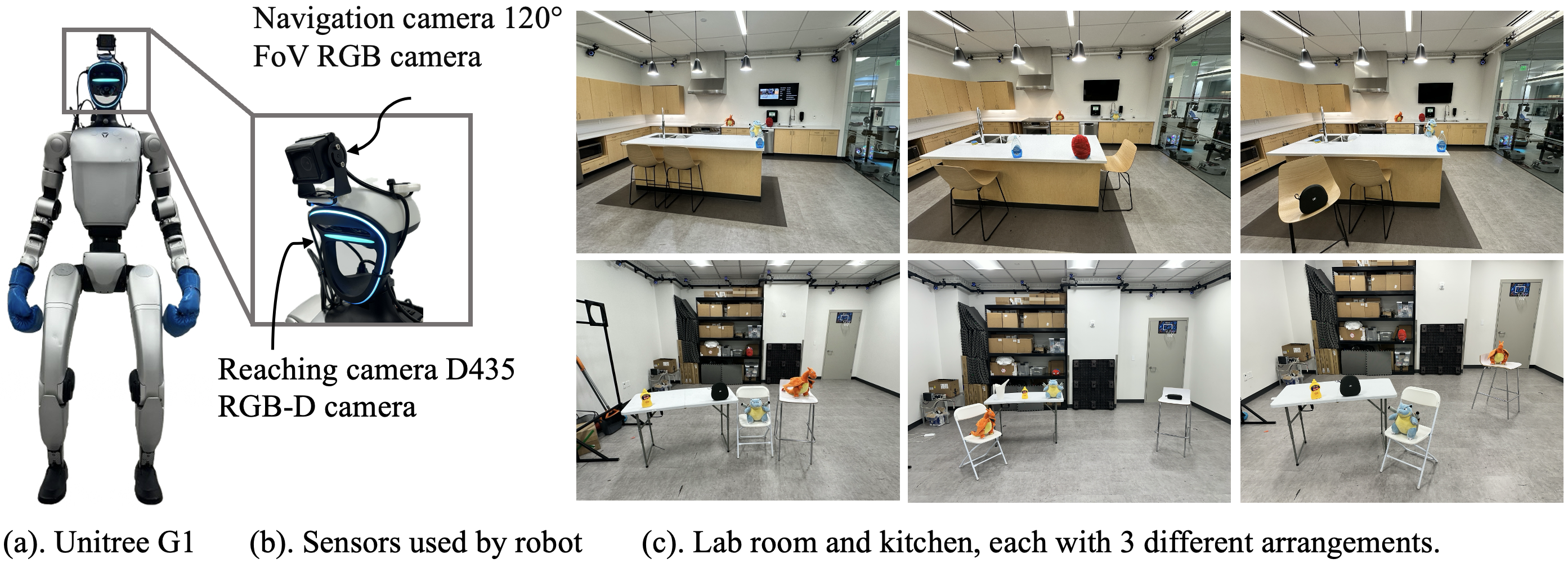}
  \vspace{-5pt}
  \caption{Hardware setup and test environments.}
  \label{fig:setup} 
\end{figure}
We build our proposed system around a Unitree G1 humanoid robot as shown in Fig.\ref{fig:setup}. For navigation, we use a wide-angle USB webcam with $90^\circ$ HFOV and $67.5^\circ$ VFOV; for the reaching module, we use G1 built-in realsense D435 RGB-D camera. During deployment, all modules are running on a PC equipped with an RTX4090 GPU and an i9-14900K CPU, and the robot is controlled via an Ethernet connection. 

We conduct our experiment in two rooms: a lab (training room) and a kitchen (deploy room). Robot-specific training data is only collected in the lab room. This is to emulate a deployment scenario where no hardware is available to record robot ground truth training data.
In each room, we arrange several pieces of furniture (e.g., shelves, chairs, tables, stools) to create diverse layouts. All test layouts and objects are unseen. 

\subsection{Whole Body Reaching with Different Scenes}
\label{sec:overall_exp}
We evaluate our method across three different layouts in each room. For each layout, robot will be ask to reach 4 objects placed at different locations and of different heights, as shown in Fig.~\ref {fig:setup}.c. Detailed experiment result is shown in Tab~\ref{tab:real_eval}. Overall, our method achieves 71 \% success rate across different environments. Success rate in lab room is 25\% greater than in the kitchen, which has narrower corridors and more reflective surfaces, which pose challenges to the goal tracker, navigation transformer and transition module. For failure cases, motion blur from robot movement may cause the tracker to lose track of the object or track the wrong object, which may confuse the navigation module. Also, humans tend to move faster and take a more aggressive path when collecting data, which may not be feasible for robots and result in a collision.
\begin{table}[h]
\centering
\begin{tabular}{llccc}
\toprule
Room                      & Scene   & Success rate & Number of misses & Number of collision \\ \midrule
\multirow{3}{*}{\bf{Lab room}} & Scene 1 & 3/4         & 1/4                & 0/4                   \\
                          & Scene 2 & 3/4         & 1/4                & 0/4                   \\
                          & Scene 3 & 4/4         & 0/4                & 0/4                   \\ \midrule
\multirow{3}{*}{\bf{Kitchen}}  & Scene 1 & 2/4            & 1/4                & 1/4                   \\
                          & Scene 2 & 2/4            & 1/4                & 1/4                   \\
                          & Scene 3 & 3/4            & 0/4                & 1/4                   \\
\bottomrule
\end{tabular}
\vspace{5pt}
\caption{Number of successes and different kinds of failures across different evaluations}
\label{tab:real_eval}
\end{table}

\subsection{Performance of Whole Body controller}
\label{sec:wbc_exp}
\textbf{Setup. } We train our whole-body controller in Isaac Gym and report its performance in two simulations: Isaac Gym~\cite{makoviychuk2021isaac} and MuJoCo~\cite{todorov2012mujoco}.
As the policy was trained in Isaac Gym, the performance reported on the left side of Table \ref{tab:sim_eval} highlights the impacts of the design choice in the training procedure. Furthermore, on the right side of Table \ref{tab:sim_eval}, the simulation-to-simulation evaluation in MuJoCo further quantifies the robustness of each policy when the simulated contact model is closer to the real world ~\cite{zhang2025wholebodymodelpredictivecontrollegged}. 



\begin{table}[]
\begin{tabular}{@{}l ccc ccc@{}}
\toprule
\multirow{3}{*}{Design Choice}                      & \multicolumn{3}{c}{Isaac Gym - Unseen Motions} & \multicolumn{3}{c}{MuJoCo - Sim2Sim Transfer} \\  
\cmidrule(lr){2-4} \cmidrule(lr){5-7}             
 &
  \begin{tabular}[c]{@{}c@{}}Pos Error\\ {[}m{]} $\downarrow$ \end{tabular} &
  \begin{tabular}[c]{@{}c@{}}Quat Error\\ {[}rad{]} $\downarrow$ \end{tabular} &
  \begin{tabular}[c]{@{}c@{}}Failure\\ {[}\%{]} $\downarrow$ \end{tabular} &
  \begin{tabular}[c]{@{}c@{}}Pos Error\\ {[}m{]} $\downarrow$ \end{tabular} &
  \begin{tabular}[c]{@{}c@{}}Quat Error\\ {[}rad{]} $\downarrow$ \end{tabular} &
  \begin{tabular}[c]{@{}c@{}}Failure\\ {[}\%{]} $\downarrow$ \end{tabular} \\ \midrule

  Ours  & \textbf{0.075}           & \textbf{0.120}          & \textbf{0}            & \textbf{0.153}          & \textbf{0.326}         & \textbf{3}          \\

  
Single discriminator              & 0.149           & 0.169          & 0          & 0.525          & 1.015           & 13          \\
No discriminator                & 0.540           & 1.138            & 97          & 1.127          & 2.044          & 99          \\

\bottomrule
\end{tabular}
\vspace{0.1cm}
\caption{Tracking accuracy is reported as positional error (m) and orientation error (rad). A timestep is counted as a failure when the head height deviates from the target by $\geq 0.4$ m. Each configuration was trained for 50 k epochs, evaluated on 1-minute unseen motion clips, and repeated five times.}

\label{tab:sim_eval}
\end{table}

\textbf{Single-stage RL Training Recipe.}  
We observed that guiding reinforcement learning (RL) exploration with generative adversarial networks (GANs) greatly improves sample efficiency. Compared to a single discriminator that jointly criticizes whole-body motion, we find that disentangling upper-body and lower-body motion via separate reward from two discriminators helps. It is probably because separate discriminators prevent the policy from entangling irrelevant motions during training and achieve a lower tracking error. 
For instance, in most walking clips, people naturally swing their arms, while dual-arm manipulation typically occurs from a squatting posture. A policy trained with a \emph{single} full-body discriminator tends to memorize the arm-swing pattern and then fails to handle a carried box while walking. In contrast, our setup decouples arm manipulation from balance control, enabling the whole-body controller to produce more diverse motions under 3-point tracking. In unseen tasks, our policy consistently outperforms the single whole-body discriminator variant.


\subsection{Performance of Navigation Module}
\label{sec:nav_exp}
\textbf{Setup. }
We use Aria Glasses to collect 200 human clips (H-Lab/H-Kit) per room. For robot data (R), we use mocap system to collect 38 clips of robot trajecotries in the lab and 20 in the kitchen. Each clip lasts around 4 seconds (human) or 30 seconds (robot). Of the lab robot data, 24 clips are used for training and 14 for testing; all kitchen robot clips are used for testing. 
This is to mimick a deployment scenario where no device is available to record robot ground truth training data. Human goals are approximated via eye gaze (Sec.\ref{sec:nav}), while robot goals are manually annotated and tracked. We also include the out-of-distribution ADT dataset (O)\cite{adt}, with 400 minutes of Aria Glasses footage of users doing tasks like cleaning and cooking.

\begin{table}[h]
\centering
\footnotesize
\begin{tabular}{l lcc  lcc}
\toprule
\multirow{2}{*}{Arch} & \multicolumn{3}{c}{Lab Room} & \multicolumn{3}{c}{Kitchen (Deploy Room)} \\
\cmidrule(lr){2-4} \cmidrule(lr){5-7}
& Training Data & SR & Error & Training Data & SR & Error \\
\midrule
shared    & H-Lab                        & 0.14 & 0.704 & --       & -- & -- \\
shared    & R-Lab                        & 0.71 & 0.427 & zero-shot        & 0.35 & 0.827 \\
shared    & R-Lab + O                    & 0.79 & 0.399 & zero-shot        & 0.35 & 0.726 \\
shared    & R-Lab + H-Lab                & 0.79 & 0.374 & + H-Kit          & 0.45 & 0.664 \\
2-branch  & R-Lab + H-Lab + O            & 0.79 & \textbf{0.356} & + H-Kit        & 0.20 & 0.812 \\
shared    & R-Lab + H-Lab + O (Ours)     & \textbf{0.86} & 0.380 & + H-Kit  & \textbf{0.60} & \textbf{0.608} \\
\bottomrule
\end{tabular}
\vspace{5pt}
\caption{\textbf{Navigation Evaluation:}  we report success rate (SR) and mean position error (Error) in open-loop prediction. }
\label{tab:nav}
\vspace{-10pt}
\end{table}

\textbf{Metrics}: We report open-loop prediction performance in both the lab room and the kitchen. For each test video, we predict a 10-step trajectory at every time step and compute the mean error against the ground-truth trajectory after rolling out these prediction. A prediction is considered successful if the final error is within 0.6 meters, which corresponds roughly to the humanoid's effective manipulation range.

\textbf{Recipe of Combining Human and Robot Data for Navigation.} As reported in Table \ref{tab:nav} Lab Room column, training with in-domain human data (H-Lab) only leads to poor success rates comparing to training with in-domain robot data (R-Lab), likely because the embodiment differences between humans and robots are not learned. The model that trains with in-domain robot data (R-Lab) together with human data, either from out-of-distribution (O) or from in-domain human data (H-Lab), further increases the performance. Human data in the same environment (H-Lab) helps moderately more than out-of-distribution data (O). By combining all existing data, we achieve the best performance. 


\textbf{Deploying into New Scenes.} As reported in Table~\ref{tab:nav} Kitchen column, while training with only robot data shows reasonable performance in the lab room, it fails to generalize to the new (deploy) room.  Collecting additional in-domain human data (+H-Kit) alone helps the robot to better in the new scenes.  Note that although incorporating ADT (O) does not significantly improve performance within the lab room, it substantially boosts success rates in the deploy room, where no robot training data is available. This highlights the importance of leveraging diverse, unlabeled human data to enhance cross-scene generalization.

\textbf{Shared Decoder Branch Improves Navigation Generalization.}
In contrast to common practice in manipulation tasks\cite{egomimic,human_humanoid}, we find that sharing a single decoding branch for human and robot data improves scene generalization in navigation tasks. We hypothesize that this is because the embodiment differences between humans and robots in short-term navigation are smaller than those in manipulation, making a shared representation more effective.

\section{Conclusion}
\label{sec:conclusion}
We presented \ours, an autonomous hand-eye delivery system for humanoid navigation and reaching. Our method achieves a 71\% success rate in reaching different objects placed in two different environments with obstacles. A future extension could be building a general grasp framework that can grasp different objects placed at various locations. Learning more fine-grained whole body navigation that can be aware and avoid collision with varying parts of the body will also be an interesting direction for a humanoid robot to be useful in real-life environments.
\newpage
\section{Limitations}
Although our method can work in environments with obstacles, it only collects human head poses and uses it as a robot control interface without considering other parts of the body. For reaching the heavily occluded target, human will utilize their whole body coordination to avoid collision, such as going sideways when facing a narrow gap or stepping over lower obstacles. Using our method will cause the robot to take a more conservative approach when facing a complex environment without fully utilizing its agility. Moreover, using head and wrist poses as a humanoid control interface has its intrinsic ambiguity for controlling lower body and may cause the humanoid robot to hesitate. For example, when the three-point is moving forward, it is hard for a humanoid to know whether the high-level intention is to bend forward or walk forward. More information that could be estimated from egocentric devices such as foot pose may help to reduce hesitation when taking actions.

\clearpage
\acknowledgments{}


\bibliography{reference}  
\newpage
\appendix
\renewcommand{\thefigure}{S\arabic{figure}}
\renewcommand{\thetable}{S\arabic{table}}
\def\theequation{S\arabic{equation}}
\setcounter{figure}{0}
\setcounter{table}{0}
\setcounter{equation}{0}
\setcounter{page}{1}

\section{Whole Body Controller}

\begin{figure}
    \centering
    \includegraphics[width=\linewidth]{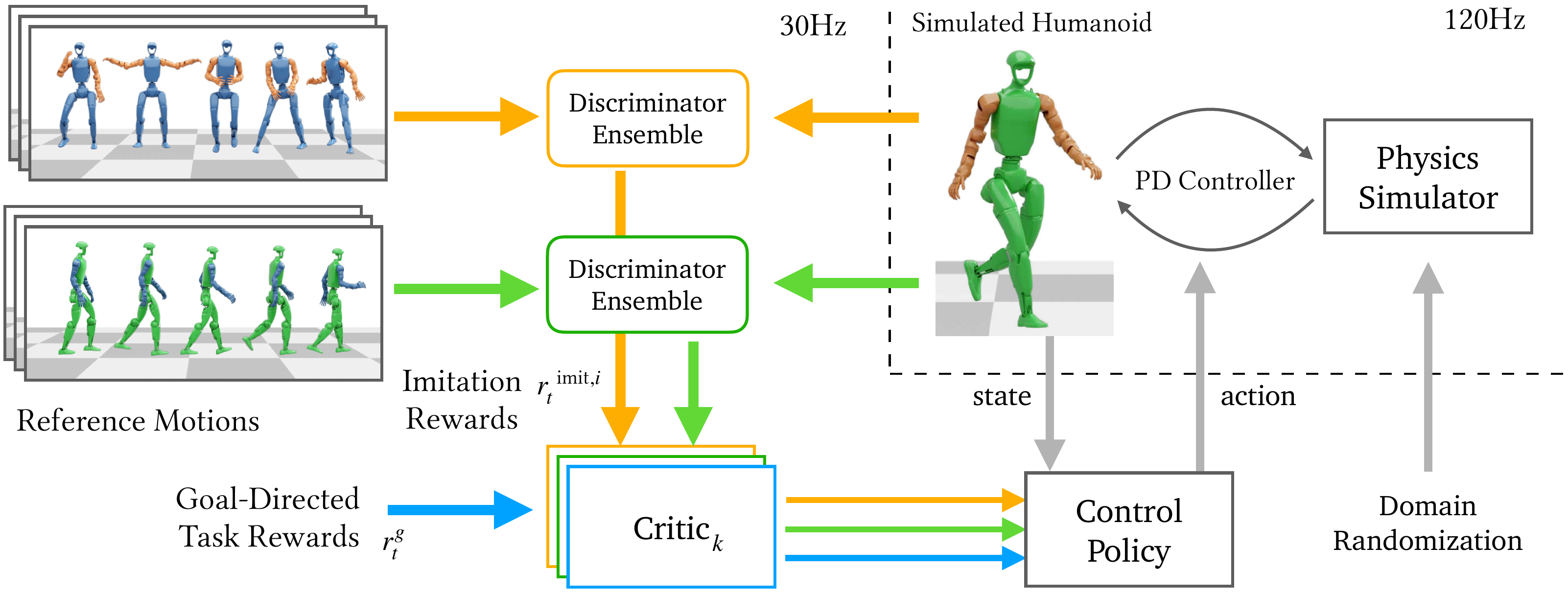}
    \caption{Systemic overview of the training scheme of our whole-body controller. We employ a multi-objective learning framework plus a GAN-like architecture for imitation learning. Our scheme allows imitating arm poses (orange) and those of the rest body parts (green) from different sources of reference motions simultaneously, and uses a goal-directed reward to fulfill the sparse tracking task for the head and hand poses.}
    \label{fig:wbc_overview}
\end{figure}

\begin{table}
\centering
\caption{Hyperparameters}
\begin{tabular}{lc}
    \toprule
    \textbf{Parameter} & \textbf{Value}\\
    \midrule
    policy network learning rate & $5 \times 10^{-6}$\\
    critic network learning rate & $1 \times 10^{-4}$\\
    discriminator learning rate & $1 \times 10^{-5}$\\
    reward discount factor ($\gamma$) & $0.95$ \\
    GAE discount factor ($\lambda$) & $0.95$ \\
    surrogate clip range ($\epsilon$) & $0.2$ \\
    gradient penalty coefficient ($\lambda^{GP}$) & $10$ \\
    number of PPO workers (simulation instances) & $1024$ \\
    PPO replay buffer size & $1024 \times 8$ \\
    PPO batch size & $256$ \\
    PPO optimization epochs & $5$ \\
    discriminator replay buffer size & $1024 \times 8 \times 2$ \\
    discriminator batch size & $512$ \\
  \bottomrule
\end{tabular}
\label{tab:hyper}
\end{table}

Figure~\ref{fig:wbc_overview} shows the overview of the systemic architecture of the whole body controller.
We use IsaacGym~\cite{makoviychuk2021isaac} as the physics engine for policy training. The policy runs at 30Hz and controls the humanoid through a PD servo.

Instead of directly imitating full-body motions, based on the motion decoupling scheme from previous literature~\cite{composite},
we split the full-body motions into arm and torso groups,
and employ two discriminators at the same time to evaluate the imitation performance for partial motions.
Besides the current state of the humanoid,
the control policy takes only the tracking target for the head and two hands as the goal input. 
Without needing the whole trajectory of full-body tracking,
the discriminators evaluate the imitation performance of partial motions and allow the arm motions to be combined with the torso and lower-body motions from different motion clips in a free way.

Given an additional goal-directed reward for target tracking plus regularizations for sim-to-real consideration,
we leverage the multi-objective learning framework~\cite{composite} to balance the learning of multiple imitation and goal-directed objectives.
The final optimization objective for policy training can be written as
\begin{equation}\label{eq:low_level_policy_optimization}
    \max \mathbb{E}_t\left[\sum\nolimits_\kappa w_k \bar{A}_{t,k} \log \pi(\mathbf{a}_t | \mathbf{o}_t) \right]
\end{equation}
where $\bar{A}_{t,k}$ is the standardized advantage that is estimated according to the achieved reward of each objective $k$, $w_k$ is an associated weight, and $\mathbf{o}_t$ is the observation including the humanoid's state $\mathbf{s}_t$ and the goal state $\mathbf{g}_t$.
We choose $w_{\text{imit}, i} = 0.2$ for each of the two imitation objectives and $w_g = 0.6$ for the goal-directed objective. 

\begin{wrapfigure}{r}{0.6\textwidth}
\vspace{-35pt}
    \centering
    \begin{subfigure}[t]{.3\linewidth}
    \includegraphics[width=\linewidth]{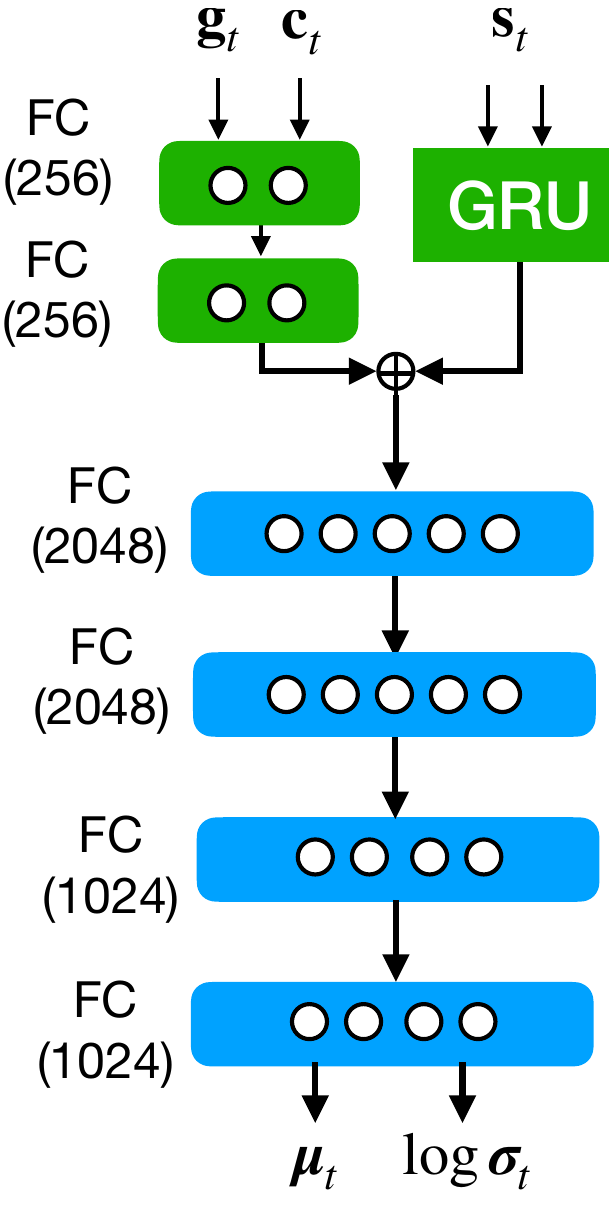}
    \caption{Policy Network}
    \end{subfigure}\quad
    \begin{subfigure}[t]{.3\linewidth}\centering
    \includegraphics[width=\linewidth]{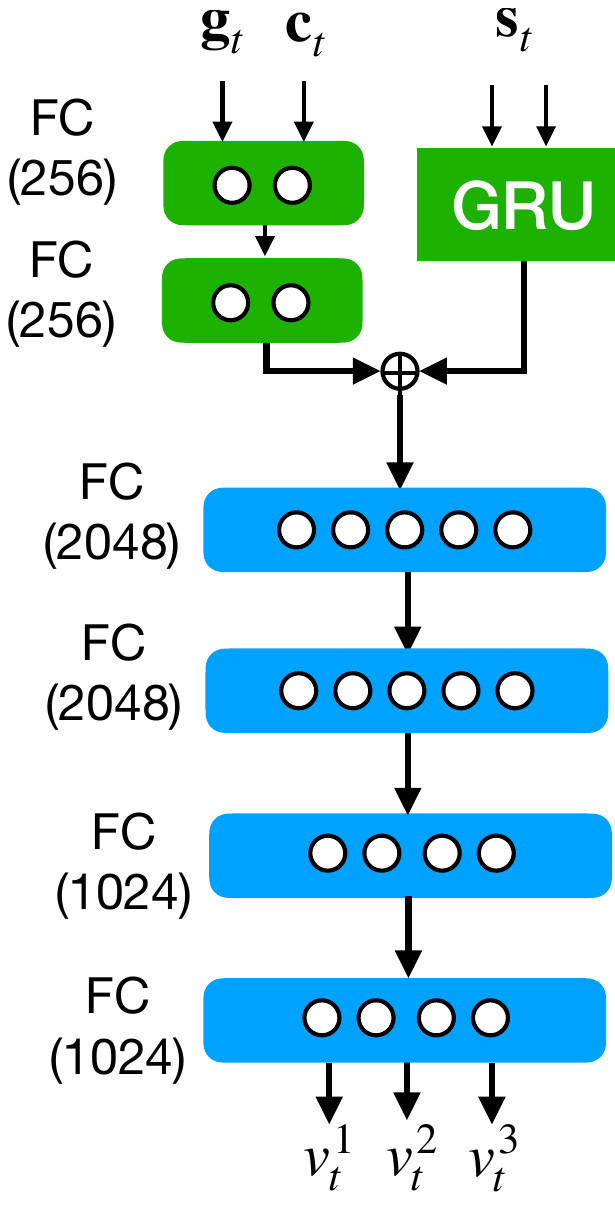}
    \caption{Value Network}
    \end{subfigure}
    \begin{subfigure}[t]{.3\linewidth}\centering
    \includegraphics[width=.9\linewidth]{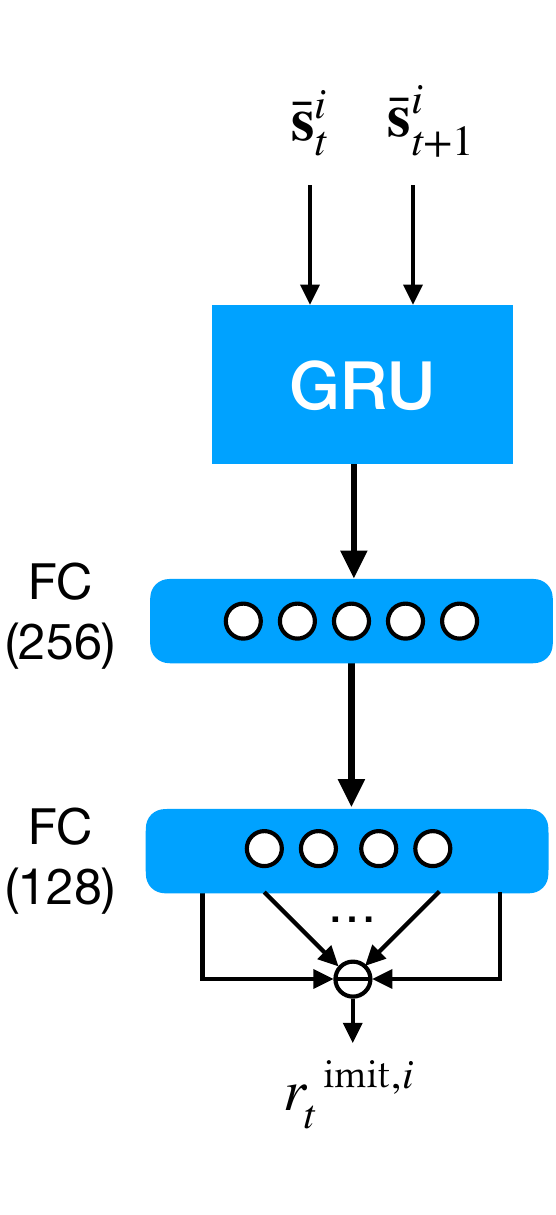}
    \caption{Discriminator}
    \end{subfigure}
    \caption{Network structures. We use $\oplus$ denoting the add operator and $\ominus$ denoting the average operator. For multi-objective learning, we employ a value network with 3-dimensional output for the two imitation objectives and one goal-directed objective.}
    \label{fig:network}
\end{wrapfigure}

We use PPO~\cite{schulman2017proximal} as the backbone reinforcement learning algorithm and take the Adam optimizer~\cite{kingma2014adam} to perform network optimization for policy training.
The hyperparameters used for policy training are listed in Table~\ref{tab:hyper} and the network structures are shown in Figure~\ref{fig:network}.
We manually pick 1363 clips of locomotion and loc-manipulation motions from AMASS~\cite{amass} and OMOMO~\cite{omomo}.
The whole data set of motions is around 5 hours long.

\subsection{Observation Space}
The G1 humanoid has 33 links and 27 controllable joints,
where the wrist and neck joints are fixed and the waist part only has 1 degree of freedom around the yaw axis.
We take two historical frames as the input and process the state vector via a GRU~\cite{cho2014learning}. This leads to a state space of $\mathbf{s}_t \in \mathbb{R}^{33 \times 7 \times 2}$ including the position and orientation (in quaternion) of each link in the local frame of the humanoid's root link, and an action space of $\mathbf{a}_t \in \mathbb{R}^{27}$.

For control purposes, we take the root angular velocity and joint velocity locally as the control state $\mathbf{c}_t \in \mathbb{R}^{30}$.
We ignore the linear velocity of the root link, since the linear velocity is hard to access from the humanoid when deployed in the real world.

The goal state vector $\mathbf{g} \in \mathbb{R}^{3\times3\times7}$ includes the target positions and orientations (in quaternion) of the three links (left hand, right hand and head) in the next three frames.

The final observation space $\mathbf{o}_t$ is composed of the three components $\mathbf{s}_t$, $\mathbf{c}_t$, and $\mathbf{g}_t$.

\subsection{Reward Terms}
Instead of using a single discriminator for each motion group,
we take the GAN-like architecture from ICCGAN~\cite{iccgan}, and employ an ensemble of 32 discriminators for each motion group.
The imitation-related reward of the discriminator ensemble $D_i$ is computed via
\begin{equation}\label{eq:dis_rew}
    r_t^{\text{imit}, i}(\mathbf{\bar{s}}_t^i, \mathbf{\bar{s}}_{t+1}^i) = \frac{1}{N} \sum_{n=1}^N \textsc{Clip}\left(D_n^i(\mathbf{\bar{s}}_t^{i}, \mathbf{\bar{s}}_{t+1}^{i}), -1, 1\right),
\end{equation}
where the subscript $i$ indicates different imitation objectives (upper or lower body groups),  $\mathbf{\bar{s}}_t^i$ is the partially observable character state for the imitation objective $i$, and the discriminator ensembles $N$ discriminators each which is trained using hinge loss~\cite{lim2017geometric} with gradient penalty~\cite{gulrajani2017improved}.
We choose $N=32$ in our implementation.

The goal-directed reward mainly measures tracking errors and also consists of three terms to stabilize the motions for sim-to-real consideration: 
\begin{equation}
    r_t^g = r_\text{tracking} + 0.8r_\text{in-air} + 0.5r_\text{sliding} + 0.005r_\text{energy}.
\end{equation}
For simplicity, here we omit the subscript $t$ for each of the reward terms.

$r_\text{tracking}$ is the reward measuring the tracking error: 
\begin{equation}
    r_\text{tracking} = 0.5\exp\left(-\frac{5}{|\mathcal{I}|}\sum_{i\in\mathcal{I}} |\varepsilon_{\text{pos},i}|\right) + 0.5\exp\left(-\frac{2}{|\mathcal{I}|}\sum_{i\in\mathcal{I}} |\varepsilon_{\text{orient},i}|\right)
\end{equation}
where $\mathcal{I} = \{\text{left hand}, \text{right hand}, \text{head}\}$ is the set of links under tracking,
$\varepsilon_{\text{pos},i}$ is the position error between the current position of the link $i$ and its target position measured in Euclidean distance,
and $\varepsilon_{\text{orient},i}$ is the orientation error measured by the angle between the orientation of the link $i$ and the target.

$r_\text{in-air} = \min\{0, t_\text{in-air}-0.5\}$ encourages the policy to keep the foot in the air for at least 0.5s during stepping.
It is computed for the swing foot when it contacts the ground,
and $t_\text{in-air}$ is the hanging time of that foot before the contact.

$r_\text{sliding} = -\sum_f c_f||\mathbf{v}_f||_2$ penalizes the linear velocity of the foot link $f$ if it contacts the ground, where $c_f = 1$ or $0$ indicating the contact state of the foot $f$.

$r_\text{energy} = - \sum_j \left(0.1|\tau_j v_j| + 0.005 \tau_j^2\right)$ penalizes the energy cost for each joint $j$, where $\tau_j$ is the torque applied on joint $j$ and $v_j$ is the joint's rotation velocity.

\subsection{Domain Randomization}
Parameters for domain randomization are listed in Table~\ref{tab:param}.

When testing the sim2real transfer, we found adding simulated delay during training time essential to prevent the robot from jittering. Compared to using a random delay across all environments, which is challenging to train, we found using a constant 1-step delay on a fixed portion $\rho_\text{delay}$ of environments improves training speed, and can prevent the robot from jittering caused by the uncertainty delay during policy execution in the real world.
We choose $\rho_\text{delay}=0.5$,
which means that the action delay is applied on half of the training environments.

\begin{table}
\centering
\caption{Domain Randomization Settings}
\begin{tabular}{lc}
    \toprule
    \textbf{Parameters} & \textbf{Value}\\
    \midrule
    Base Mass (kg) & $[-3, 3]$ \\
    Body Link Friction Coefficient & $[0.5, 1.25]$ \\
    Scale on PD Servo Gain (Kp) & $[0.7, 1.3]$ \\
    Scale on PD Servo Damping (Kd) & $[0.7, 1.3]$ \\
    Action Delay Ratio ($\rho_\text{delay}$) & $ 0.5 $ \\
  \bottomrule
\end{tabular}
\label{tab:param}
\end{table}

\section{Image Augmentation of Navigation Data}
We augment human data collected by Aria Glasses to resemble the robot’s view.
First, we address image discrepancies caused by differences in capture devices. Aria Glasses use an RGB fisheye camera with a $110^\circ$ horizontal and vertical field-of-view (HFOV, VFOV), while the robot camera has a $90^\circ$ HFOV and $67.5^\circ$ VFOV. We undistort the fisheye images to obtain $I_u$, which approximates a pinhole camera view with intrinsics $K_u$.
Second, we address discrepancies in camera pose (extrinsics) due to morphological differences. Humans tend to tilt their heads—e.g., looking downward when reaching for an object—whereas the robot’s navigation camera, fixed at the top of the head, always looks forward. As a result, the same object may appear in different regions of the image despite similar camera positions. To align viewing angles, we apply a homography to the undistorted image $I_u$ to match the pitch angle. Given the robot camera’s intrinsics $K_r$ and Aria’s effective intrinsics $K_u$, we apply $H = K_r R_{\text{pitch}} R_{[\theta]}^T K_u^{-1}$, where $R_{[\theta]}$ is the pitch component of the current human camera pose and $R_{\text{pitch}}$ is the desired robot pitch.

\section{Usage of motion capture system}
Although our whole-body controller and navigation model don't rely on world coordinates, in our real-world experiment, we found it hard to keep the robot standing at the exact location without drifting. Therefore, we attach mocap markers only to the robot's head and transform the camera trajectory into the world frame; we found that having closed-loop tracking in the world frame can significantly improve the accuracy of robot reaching.
\end{document}